\documentclass[letterpaper]{article} 
\usepackage{aaai24}  
\usepackage{times}  
\usepackage{helvet}  
\usepackage{courier}  
\usepackage[hyphens]{url}  
\usepackage{graphicx} 
\usepackage{natbib}  
\usepackage{caption} 
\usepackage{algorithm}
\usepackage{algorithmic}
\usepackage{amsfonts}
\usepackage{amssymb}
\usepackage{amsmath}
\usepackage{booktabs}
\usepackage{multirow}
\usepackage{makecell}
\usepackage{colortbl} 
\usepackage{xcolor}
\usepackage{mathrsfs}
\usepackage{float}
\newcommand{\ve}[1]{\mathbf{#1}} 
\newcommand{\ma}[1]{\mathrm{#1}} 
\usepackage{newfloat}
\usepackage{listings}
\DeclareCaptionStyle{ruled}{labelfont=normalfont,labelsep=colon,strut=off} 
\floatstyle{ruled}
\newfloat{listing}{tb}{lst}{}
\floatname{listing}{Listing}
\title{CRA-PCN: Point Cloud Completion with Intra- and Inter-level Cross-Resolution Transformers}
\author{
    Yi Rong,
    Haoran Zhou,
    Lixin Yuan,
    Cheng Mei,
    Jiahao Wang,
    Tong Lu\thanks{Corresponding author}
}
\affiliations{
   State Key Laboratory for Novel Software Technology, Nanjing University\\

    \{rongyi, dg1833031, meicheng, wangjh\}@smail.nju.edu.cn,
    hrzhou98@gmail.com,
    lutong@nju.edu.cn\\

}

\usepackage{bibentry}

\begin{document}

\maketitle

\begin{abstract}
Point cloud completion is an indispensable task for recovering complete point clouds due to incompleteness caused by occlusion, limited sensor resolution, etc. 
The family of coarse-to-fine generation architectures has recently exhibited great success in point cloud completion and gradually became mainstream.  
In this work, we unveil one of the key ingredients behind these methods: meticulously devised feature extraction operations with explicit cross-resolution aggregation.
We present Cross-Resolution Transformer that efficiently performs cross-resolution aggregation with local attention mechanisms. 
With the help of our recursive designs, the proposed operation  can capture more scales of features than common aggregation operations, which is beneficial for capturing fine geometric characteristics.  
While prior methodologies have ventured into various manifestations of inter-level cross-resolution aggregation, the effectiveness of intra-level one and their combination has not been analyzed. 
With unified designs, Cross-Resolution Transformer can perform intra- or inter-level cross-resolution aggregation by switching inputs. 
We integrate two forms of Cross-Resolution Transformers into one up-sampling block for point generation, and following the coarse-to-fine manner, we construct CRA-PCN to incrementally predict complete shapes with stacked up-sampling blocks. 
Extensive experiments  demonstrate that our method outperforms state-of-the-art methods by a large margin on several widely used benchmarks. 
Codes are available at \url{https://github.com/EasyRy/CRA-PCN}.
\end{abstract}

\section{Introduction}

\begin{figure}[t]
\centering 
\includegraphics[width=1.0\linewidth]{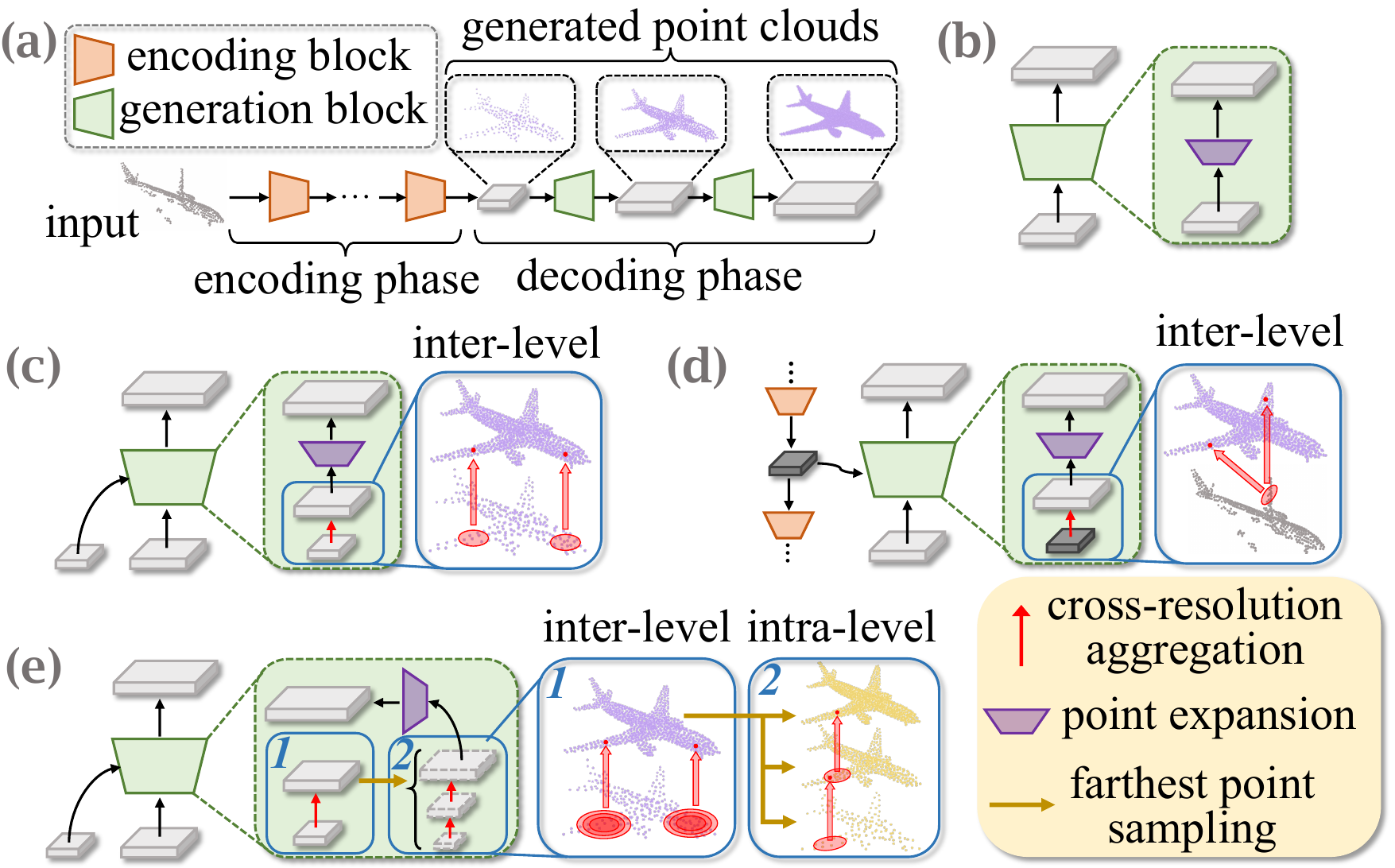}
\caption{
Illustration of our main idea.
Here, we analyze several point generation methods from the perspective of {\em cross-resolution aggregation} (CRA). 
(a) Common pipeline of coarse-to-fine completion approaches. 
(b) The plain generation operation simply generates points without considering explicit CRA.  
(c) \& (d) Several methods exploit skip connections to aggregate features of other generated point clouds or partial inputs for the current point cloud, which can efficiently  capture multi-scale features. 
(e) Our method not only extracts more fruitful multi-scale features with novel-designed enhanced inter-level CRA but also combines intra-and inter-level CRA for better capturing geometric characteristics.}
\label{fig:intro}
\end{figure}

Driven by the  rapid development of 3D  acquisition technologies, 3D vision is in great demand for research. 
Among various types of 3D data, point cloud is the most  popular description and commonly used in real-world applications~\cite{cadena2016past, reddy2018carfusion, rusu2008towards}. 
However, due to self-occlusion and limited sensor resolution, acquired point clouds are usually highly sparse and incomplete, which impedes further applications. 
Therefore, recovering complete point clouds is an indispensable task, whose major purposes are to preserve details of partial observations, infer missing parts, and densify sparse surfaces.

In the realm of deep learning, various approaches have been proposed to tackle this problem~\cite{yang2018foldingnet, choy20163d, girdhar2016learning, xie2020grnet}. 
Especially with the success of PointNet~\cite{qi2017pointnet} and its successors~\cite{qi2017pointnet++,wang2019dynamic,zhao2021point}, most approaches recover complete point clouds directly based on  3D coordinates~\cite{yuan2018pcn, wen2020point, wen2021pmp, tchapmi2019topnet, yu2021pointr, wang2020cascaded}. 
Due to the unordered and unstructured nature of point cloud data, learning fine geometric characteristics and structural features is essential for predicting reasonable shapes. 
To accomplish this aim, mainstreams of the recent methods~\cite{xiang2021snowflakenet, zhou2022seedformer, huang2020pf, yan2022fbnet, huang2020pf, tang2022lake, wang2022learning} formulate the point cloud completion task as a tree-like generation process where an encoder-decoder architecture is  adopted to extract shape representation ({\em e.g.}, a latent vector representing a complete shape) and subsequently recover the complete point cloud from low-resolution to high-resolution, as shown in Fig.~\ref{fig:intro}(a). 

As illustrated in Fig.~\ref{fig:intro},  we select several representative methods and conduct an analysis from the perspective of {\em cross-resolution aggregation} (CRA), which is a crucial component of these hierarchical methods. 
Prior to generating the points, it is crucial to extract semantic information from the current point cloud.
The plain methods~\cite{yuan2018pcn, huang2020pf} adopt common feature extraction operations without considering aggregating features from point clouds with different resolutions, namely cross-resolution aggregation. 
Based on skip connections (or dense connections) among different generation stages, the methods showed in Fig.~\ref{fig:intro}(c) realize cross-resolution aggregation via interpolation followed by multi-layer perceptrons~\cite{yifan2019patch} or transformers~\cite{yan2022fbnet,  zhou2022seedformer}. 
As illustrated in Fig.~\ref{fig:intro}(d), U-Net-like methods~\cite{wen2020point, yu2021pointr} aggregate  features of partial inputs for current point clouds, commonly based on global attention mechanisms due to the mismatching of shapes, yet another form of cross-resolution aggregation.
While plain methods can extract multi-scale representations in the generation phase, the success of  the latter two methods suggests that explicit cross-resolution aggregation is beneficial for efficiently capturing multi-scale features and  promoting the completion performance. 

Although prior approaches accomplished cross-resolution aggregation with inter-layer connections ({\em i.e.}, inter-level CRA), they overlooked cross-resolution aggregation inside layers ({\em i.e.}, intra-level CRA).
In the process of point cloud completion, missing points are not generated all at once but are gradually completed through layer-by-layer generation, which indicates that the details of the intermediate point clouds are not exactly the same, making it hard for inter-level CRA-based methods to find generation patterns that fit local regions best. 
In other words, it is crucial to combine inter-level and intra-level CRA, as shown in Fig.~\ref{fig:intro}(e), to overcome the flaws of previous methods. 

To this end, we propose Cross-Resolution Transformer that performs multiple scales of cross-resolution aggregation with recursive designs and local attention mechanisms. 
Cross-Resolution Transformer has several favorable properties: 
(1) {\em Explicit cross-resolution aggregation}: Point cloud directly aggregates features from ones with different resolutions without intermediate aids; 
(2) {\em Multi-scale aggregation}: Unlike previous methods, which only realized CRA  on a few scales, our method can extract more scales of features with recursive designs, even taking two resolutions of point clouds as input; 
(3) {\em Unified design}: Intra- and inter-level CRA  share the same implementation, and different forms of CRA can be achieved by switching inputs; 
(4) {\em Plug-and-play}: Without introducing significant computation, Cross-resolution Transformer can serve as a plug-and-play module to extract local features in the decoder or encoder with large receptive fields.

We integrate  two  forms of Cross-Resolution Transformers  in one up-sampling block  for intra- and inter-level cross-resolution aggregation, respectively.
Thanks to this combination, the decoding block can precisely capture multi-scale geometric characteristics and generate points that best fit local regions. 
Mainly based on the up-sampling block, we propose CRA-PCN for point cloud completion. 
CRA-PCN has an encode-decoder architecture, and its decoder consists of three consecutive up-sampling blocks. 
With the help of this effective decoder, our method outperforms state-of-the-art completion networks on several widely used benchmarks. 
The contributions in this work can be summarized as follows:

\begin{itemize}
\item We show that one of the key ingredients behind the success of prior methods is explicit cross-resolution aggregation (CRA) and propose the combination of inter- and intra-level CRA to extract fine local features.
\item We devise an effective local aggregation operation named Cross-Resolution Transformer, which can adaptively summarize multi-scale  geometric characteristics in the manner of cross-resolution aggregation. 
\item We propose a novel CRA-PCN for point cloud completion, which precisely captures multi-scale local properties  and predicts  rich details.
\end{itemize}

\begin{figure*}[t]
\begin{center}
\includegraphics[width=1.0\linewidth]{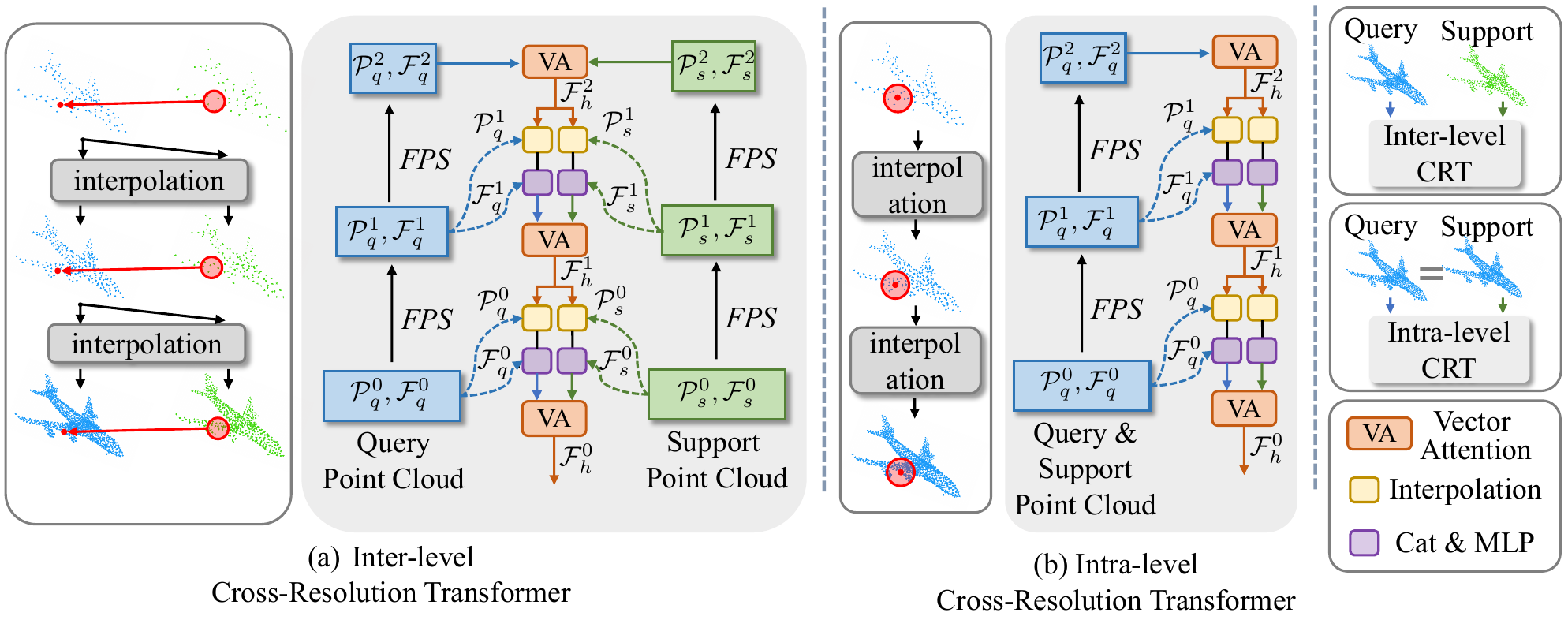}
\end{center}
\caption{
Illustration of Cross-Resolution Transformer (CRT).
CRT considers the cross-resolution aggregation on $m$ scales, and $m=3$ in this figure. 
(a) Inter-level CRT lets the query point cloud (blue) aggregate features from the support one (green); both of them are intermediate point clouds during the generation phase. 
(b) Intra-level CRT realizes cross-resolution aggregation inside the current point cloud and is a degenerate form of inter-level one.}
\label{fig:crt}
\end{figure*}

\section{Related Work} \label{sec:rw}

\subsection{Point Cloud Learning}

Early works usually adopt multi-view projection~\cite{li2016vehicle,chen2017multi,lang2019pointpillars} 
or 3D voxelization~\cite{maturana2015voxnet,song2017semantic,riegler2017octnet,choy20194d} to transform the irregular point clouds into regular representations, followed by 2D/3D CNNs. 
However, the converting costs are expensive, and geometric details will inevitably be lost in the transforming process.
Therefore, researchers have designed deep networks which can directly process 3D coordinates based on permutation-invariant operators. 
PointNet~\cite{qi2017pointnet} and PointNet++~\cite{qi2017pointnet++} are the pioneering point-based networks which adopt MLPs and max pooling to extract and aggregate features across the whole set or around local regions. 
Based on them, a number of point-based methods have been proposed. 
Some methods~\cite{wang2019dynamic, zhao2019pointweb,simonovsky2017dynamic} convert the local region to a graph, followed by graph convolution layers.
Other methods~\cite{xu2018spidercnn, wu2019pointconv,thomas2019kpconv} define novel convolutional operators that apply directly to the 3D coordinates without quantization. 
Recently, attention-based methods, especially the family of transformers~\cite{vaswani2017attention, zhao2021point, guo2021pct,park2022fast}, have achieved impressive success thanks to the inherent permutation-invariant, where the whole point set or local region is converted to a sequence and the aggregation weights are adaptively 
determined by data.
Among them, we exploit vector attention mechanisms~\cite{zhao2021point} to construct our Cross-Resolution Transformer.

\subsection{Point Cloud Completion}
Like traditional methods for point cloud learning, early attempts~\cite{choy20163d, girdhar2016learning, han2017high} at 3D shape completion usually adopt intermediate aids ({\em i.e.}, voxel grids). 
However, these methods usually suffer from heavy computational costs and geometric information loss.
Boosted by the point-based learning mechanisms discussed above, researchers pay more attention to designing point-based completion methods. 
PCN~\cite{yuan2018pcn} is the first learning-based work that adopts an encoder-decoder architecture.
It recovers point cloud in a two-stage process where a coarse result is predicted by MLP and then the fine result is predicted with folding operation~\cite{yang2018foldingnet} from the coarse one.
Along this line, more methods~\cite{xiang2021snowflakenet, zhou2022seedformer,yan2022fbnet,tang2022lake,wang2022learning, li2023proxyformer, chen2023anchorformer} spring up and 
yield impressive results with the help of  more generation stages, better feature extraction, or structured 
generation process.
As discussed before, explicit cross-resolution aggregation is another key ingredient behind their success.
Following the coarse-to-fine manner, we propose CRA-PCN, and our insight into promoting point cloud completion is to introduce well-designed cross-resolution aggregation mechanisms.

\section {Method}
In this section, we will first elaborate on the details of Cross-Resolution Transformer. 
Then, we will illustrate the architecture of CRA-PCN in detail.
Lastly, we will introduce the loss function used for training.

\begin{figure*}[t]
\begin{center}
\includegraphics[width=1.0\linewidth]{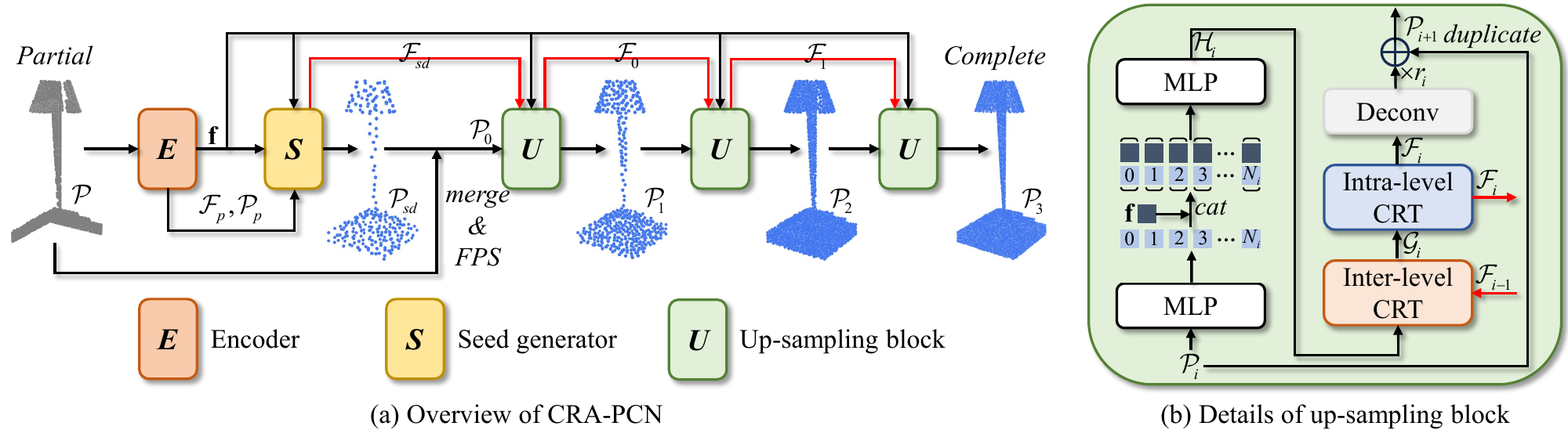}
\end{center}
\caption{
(a) The overall architecture of CRA-PCN, which consists of encoder, seed generator, and stacked up-sampling blocks. 
(b) The details of the up-sampling block, which is composed of MLP, inter-level Cross-Resolution Transformer, intra-level Cross-Resolution Transformer, and deconvolution.}
\label{fig:arc}
\end{figure*}

\subsection{Inter-level Cross-Resolution Transformer}
As shown in Fig.~\ref{fig:crt}(a), the purpose of Cross-Resolution Transformer (CRT for short) is to aggregate features of the support point cloud (green) for the query point cloud (blue) on multiple scales. 
The query point cloud is the current point cloud to be up-sampled, and the support one was generated in the early generation stage; therefore, the query one has a higher resolution than the support one. 
We design CRT in a  recursive manner, which means it can work on arbitrary-number scales and the computational complexity is bounded.

Given coordinates $\mathcal{P}_q \in {\mathbb R}^{N_q \times 3}$ and the corresponding features $\mathcal{F}_q  \in {\mathbb R}^{N_q \times D}$ of the query point cloud with $\mathcal{P}_s \in {\mathbb R}^{N_s \times 3}$ and  $\mathcal{F}_s  \in {\mathbb R}^{N_s \times D}$ of the support one, 
we first adopt hierarchical down-sampling to obtain their subsets, respectively. 
The number of subsets, namely number of scales, is $m$. 
The subsets of original  coordinates and features are denoted as $\{\mathcal{P}_q^l\}_{l=0}^{m-1}$, $\{\mathcal{F}_q^l\}_{l=0}^{m-1}$, $\{\mathcal{P}_s^l\}_{l=0}^{m-1}$, and $\{\mathcal{F}_s^l\}_{l=0}^{m-1}$, like shown in Fig.~\ref{fig:crt}(a). 
Note that the index $0$ corresponds to the original point cloud. 
Then, we exploit the local attention mechanism~\cite{zhao2021point} to achieve cross-resolution aggregation on the $(m-1)$-th scale. 
The attention relations in the neighborhood of each query point can be calculated as follows:

\begin{equation}
{\hat{\ve{a}} ^{m-1}_{ij}} = {\alpha^{m-1}} (\ve{q}^{m-1}_i - \ve{k}^{m-1}_j+\delta^{m-1}_{ij}),
\end{equation}

{\noindent}where ${\alpha^{m-1}}$  is a non-linear projection function implemented with multi-layer perceptron (MLP).
Let $\ve{p}^{m-1}_{q, i}$ be the $i$-th coordinate of $\mathcal{P}_q^{m-1}$, we use $k\rm nn$ algorithm to search the indexes of its neighborhood in the support point cloud, which is denoted as ${\rm idx}(\ve{p}^{m-1}_{q, i}) = \{j \in k{\rm nn}(\ve{p}^{m-1}_{q, i})\}$.
$\delta^{m-1}_{ij} \in \mathbb{R}^{D}$ is the position encoding generated by subtraction between $\ve{p}^{m-1}_{q,i}$  and $\ve{p}^{m-1}_{s, j}$ followed by a MLP.
$\ve{q}^{m-1}_i$ is the query vector projected from the $i$-th feature vector of $\mathcal{F}_q^{m-1}$ via linear projection, and $\ve{k}^{m-1}_j$ is projected from the corresponding feature vector of $\ve{p}^{m-1}_{s, j}$ ({\em i.e.}, $\ve{f}^{m-1}_{s, j}$). 
For all $j \in {\rm idx}(\ve{p}^{m-1}_{q, i})$, we normalize ${\hat{\ve{a}} ^{m-1}_{ij}}$ for ${\ve{a}} ^{m-1}_{ij}$ with channel-wise $\ma{softmax}$ function and  aggregate local features:

\begin{equation}
\ve{h}_i^{m-1}  = \sum_{j \in {\rm idx}(\ve{p}^{m-1}_{q, i}) }\ve{a}_{ij}^{m-1} \odot  (\ve{v}_j^{m-1} + \delta^{m-1}_{ij}). 
\end{equation} 

{\noindent}Here, $\ve{h}_i^{m-1}$ indicates the $i$-th feature of $\mathcal{F}_h^{m-1}$, as shown in Fig.~\ref{fig:crt}(a), and $\ve{v}^{m-1}_j$ is projected from $\ve{f}^{m-1}_{s, j}$ via linear projection, like $\ve{k}^{m-1}_j$. 
Next, we adopt feature interpolation \cite{qi2017pointnet++} to map $\mathcal{F}_h^{m-1}$ onto the coordinates $\mathcal{P}_q^{m-2}$ and $\mathcal{P}_s^{m-2}$, 
and the two interpolated features are concatenated with $\mathcal{F}_q^{m-2}$ and $\mathcal{F}_s^{m-2}$ followed by MLPs for feature fusion, respectively. 
The two enhanced features are passed through the aforementioned attention mechanism for $\mathcal{F}_h^{m-2}$. 
The process is iterated until we obtain $\mathcal{F}_h^{0}$, which is the output of CRT.

The common cross-resolution aggregation operations only work on the $0$-th scale, {\em i.e.}, the bottom scale shown in Fig.~\ref{fig:crt}. 
We argue that these single-scale designs restrict the modeling power, leading to unreasonable predictions. 
While several methods~\cite{qian2021pu, he2023grad} adopt stacked feature extraction layers with dense connections among these layers to capture multi-scale features, the computational complexities of them are not bounded as the number of layers grows, and the receptive fields grow slowly without explicit hierarchical down-sampling.

\subsection{Intra-level Cross-Resolution Transformer}
With the help of unified implementation, let the query point cloud and support one be the same, inter-level CRT degrades to the intra-level one. 
In this case, the cross-resolution aggregation is decomposed into self-attention, interpolation, and feature fusion, as illustrated in Fig.~\ref{fig:crt}(b). 

Inter-level CRT plays an important role in preserving learned features and capturing multi-scale features.
However, the generated point cloud with lower resolution is not necessarily a subset of the current one, and the local details of them are sometimes different due to  irregular outliers or shrinkages caused by unreasonable coordinate predictions. 
Therefore, only using inter-level CRTs is not sufficient for finding fine generation patterns that fit local regions best. 
To alleviate this problem, we introduce intra-level  CRT and integrate both into an up-sampling block, like shown in Fig.~\ref{fig:arc}(b). 
Note that the combination of inter- and intra-level CRT is not an incremental improvement. 
The experiments demonstrate that their combination outperforms the variants only using two consecutive intra- or inter-level CRTs by a large margin  with the same number of parameters.

\begin{table*} [t]
\begin{center}
\begin{tabular}{c |  c | cccc cccc}
\toprule
Methods & Average & Plane &Cabinet& Car& Chair& Lamp &Couch &Table  &Boat\\
\midrule
FoldingNet~\cite{yang2018foldingnet}	&14.31	&9.49	&15.80	&12.61	&15.55	&16.41	&15.97	&13.65	&14.99\\
TopNet~\cite{tchapmi2019topnet}		&12.15	&7.61	&13.31	&10.90	&13.82	&14.44	&14.78	&11.22	&11.12\\
PCN~\cite{yuan2018pcn}		&9.64	&5.50	&22.70	&10.63	&8.70	&11.00	&11.34	&11.68	&8.59\\
GRNet~\cite{xie2020grnet}	&8.83	&6.45	&10.37	&9.45	&9.41	&7.96	&10.51	&8.44	&8.04\\
PoinTr~\cite{yu2021pointr}		&8.38	&4.75	&10.47	&8.68	&9.39	&7.75	&10.93	&7.78	&7.29\\
SnowflakeNet~\cite{xiang2021snowflakenet}&7.21	&4.29	&9.16	&8.08	&7.89	&6.07	&9.23	&6.55	&6.40\\
FBNet~\cite{yan2022fbnet}&6.94	&3.99	&9.05	&7.90	&7.38	&5.82	&8.85	&6.35	&6.18\\
ProxyFormer~\cite{li2023proxyformer} & 6.77 & 4.01& 9.01& 7.88& 7.11& 5.35& 8.77& 6.03& 5.98\\ 
SeedFormer~\cite{zhou2022seedformer} &6.74	&3.85	&9.05	&8.06	&7.06	&5.21	&8.85	&6.05	&5.85\\
AnchorFormer~\cite{chen2023anchorformer} & 6.59 & 3.70 &8.94& 7.57 &7.05 &5.21& 8.40 &6.03 &5.81\\
\midrule
Ours &\bf6.39	&\bf3.59	&\bf8.70	&\bf7.50	&\bf6.70	&\bf5.06	&\bf8.24	&\bf5.72	&\bf5.64\\
\bottomrule
\end{tabular}
\end{center}
\caption{Results on PCN dataset in terms of L1 Chamfer Distance $\times$ $10^3$ (lower is better). } \label{tab:pcn1}
\end{table*}

\begin{figure*}[!t]
\begin{center}
\includegraphics[width=1.0\linewidth]{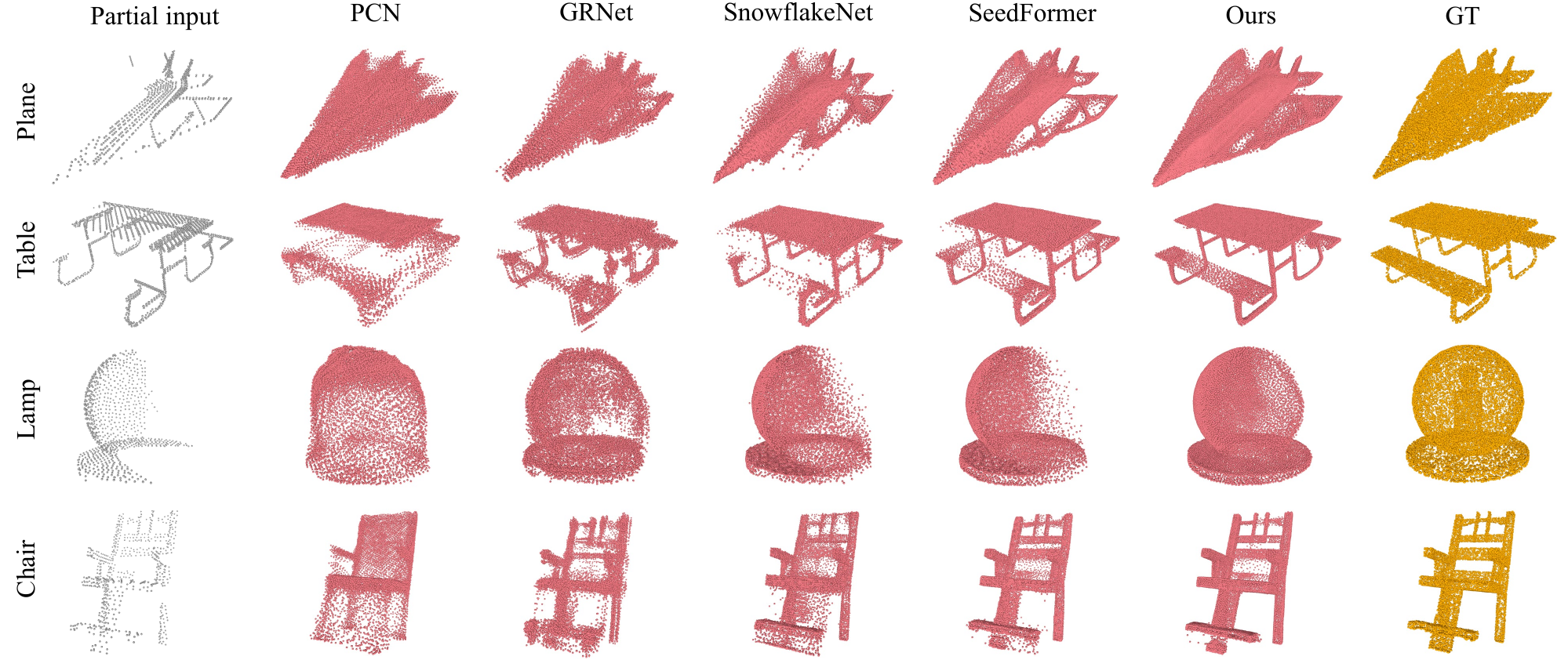}
\end{center}
   \caption{Visual comparison on PCN dataset.}
\label{fig:pcn}
\end{figure*}

\subsection{CRA-PCN}
The overall architecture of CRA-PCN is illustrated in Fig.~\ref{fig:arc}(a), which consists of encoder, seed generator, and decoder. 
The decoder consists of three up-sampling blocks.

\subsubsection{Encoder and seed generator.} 
The purpose of the encoder is to extract a shape vector and per-point features of pooled partial input.
The input of the encoder is the partial point cloud $\mathcal{P} \in \mathbb{R}^{N \times 3}$.
With hierarchical aggregation and down-sampling followed by a max-pooling operation, we can obtain the shape vector $\ve{f} \in \mathbb{R}^{C}$ and down-sampled partial points  
$\mathcal{P}_p \in \mathbb{R}^{N_p \times 3}$ and corresponding features $\mathcal{F}_p \in \mathbb{R}^{N_p \times C_p}$.
Specifically, we adopt three layers of set abstraction~\cite{qi2017pointnet++} (SA layer) to down-sample point set and aggregate local features; moreover, we insert an intra-level CRT  between two consecutive SA layers to enrich semantic contexts. 

The seed generator aims to produce a representation with coordinates and features representing a sketch point cloud with low resolution yet complete shape, which is named seed.
To leverage the impressive detail-preserving ability of Upsample Transformer~\cite{zhou2022seedformer} for seed generation, we feed  $\mathcal{P}_p$ and $\mathcal{F}_p$  into Upsample Transformer without $\rm {softmax} $ activation. 
The seed generation phase can be formulated as follows:
\begin{equation}
{\mathcal F}_{sd} = {\mathrm {UpTrans}}({\mathcal P}_p, {\mathcal F}_p), {\mathcal P}_{sd} = {\rm MLP} ({\rm cat}({\mathcal F}_{sd}, {\ve f})),
\end{equation}

{\noindent}where ${\mathcal F}_{sd} \in \mathbb{R}^{N_{sd} \times D}$ is the set of seed features that carry the local properties of seed points, and $\rm UpTrans$ indicates Upsample Transformer. 
We concatenate ${\mathcal F}_{sd}$ with $\ve{f}$ and then feed the concatenated features into a MLP for seed coordinates ${\mathcal P}_{sd} \in \mathbb{R}^{N_{sd} \times 3}$. 
For better optimization, following prior methods~\cite{xiang2021snowflakenet, wang2020cascaded}, we merge partial input $\mathcal{P}$ with ${\mathcal P}_{sd} $ and down-sample the merged set to obtain the starting points ${\mathcal P}_0\in \mathbb{R}^{N_0\times 3}$.

\subsubsection{Decoder.}
The objective of the decoder is to gradually recover the complete shape from starting points $\mathcal{P}_0$.
Our decoder consists of three up-sampling blocks.
Given starting coordinates ${\mathcal  P}_0$, the decoder will gradually produce ${\mathcal P}_1$, ${\mathcal P}_2$, and ${\mathcal P}_3$,  where ${\mathcal P}_{i} \in \mathbb{R}^{N_i \times 3}$ is the input of the $i$-th block and $N_{i+1}=N_{i} \times r_i$ for each $i<2$.

\subsubsection{Up-sampling block.}
As illustrated in Fig~\ref{fig:arc}(b), each block is mainly composed of a mini-PointNet~\cite{qi2017pointnet}, inter-level CRT, intra-level CRT, and deconvolution~\cite{xiang2021snowflakenet}. 
The $i$-th block can upsample current point set $\mathcal{P}_{i}$ at the rate of $r_i$. 
With shape vector $\ve f$,  we first adopt a mini-PointNet~\cite{qi2017pointnet} to learn per-point features $\mathcal{H}_i \in \mathbb{R}^{N_i \times D}$.
Then, let $\mathcal{P}_{i-1}$ be the coordinates of the support point cloud with corresponding features $\mathcal{F}_{i-1}$ and $\mathcal{P}_{i}$ be the coordinates of the query one with corresponding features $\mathcal{H}_{i}$,  we employ inter-level CRT to extract features ({\em i.e.}, ${{\mathcal G}_i \in {\mathbb R}^{N_i \times D}}$) in the cross-resolution manner. 
Note that, in the case of $i=0$, ${\mathcal P}_{i-1}={\mathcal P}_{sd}$ and ${\mathcal F}_{i-1}={\mathcal F}_{sd}$. 
Next, ${\mathcal G}_i$ and ${\mathcal P}_i$ are fed into intra-level CRT to capture internal multi-scale geometric features and obtain feature $\mathcal{F}_i$. Finally, with $\mathcal{F}_i$, we use deconvolution to predict $r_i$ offsets for each point and subsequently obtain up-sampled coordinates ${\mathcal P}_{i+1}$.
Note that, $\mathcal{F}_i$ and $\mathcal{P}_i$ are sent to the next up-sampling block via skip connection. 

\begin{table*} [!t]
\begin{center}
\setlength{\tabcolsep}{1mm}{
\begin{tabular}{c |ccccc | ccccc| ccc| c}
\toprule
Methods & Table & Chair &\makecell*[c]{ Air \\ plane} & Car& Sofa&  \makecell*[c]{ Bird \\ house}&Bag &Remote  & \makecell*[c]{Key\\ board}& Rocket & CD-S & CD-M & CD-H & CD-Avg\\
\midrule
FoldingNet 	&2.53	&2.81&	1.43	&1.98&	2.48	&	4.71&2.79&	1.44&	1.24	&1.48&		2.67&	2.66	&4.05&	3.12\\
PCN			&2.13	&2.29&	1.02	&1.85&	2.06	&	4.50	&2.86&	1.33&	0.89	&1.32	&	1.94&	1.96	&4.08&	2.66\\
TopNet		&2.21	&2.53&	1.14	&2.28&	2.36	&	4.83&2.93&	1.49&	0.95	&1.32&		2.26	&2.16&	4.30&	2.91\\
PFNet		&3.95	&4.24&	1.81	&2.53&	3.34	&	6.21	&4.96&	2.91	&1.29&	2.36	&	3.83&	3.87&	7.97&	5.22\\
GRNet		&1.63	&1.88&	1.02	&1.64&	1.72	&	2.97	&2.06&	1.09&0.89 &1.03	&	1.35&	1.71&	2.85&	1.97\\
PoinTr		&0.81	&0.95&	0.44	&0.91&	0.79	&	1.86 &0.93& 0.53& 0.38& 0.57	&	0.58	&0.88&	1.79	&1.09\\
SeedFormer	&0.72	&0.81&	0.40	&0.89&	0.71	&	1.51&    0.79&     0.46 &     0.36  &  	\bf{0.50}&	0.50	&0.77&	1.49&	0.92\\
\midrule
Ours &\bf{0.66	}&\bf{0.74}&	\bf{0.37} &\bf{0.85}&	\bf{0.66}&	\bf{1.36 }	&\bf{0.73}& \bf{0.43	}&\bf{0.35} &\bf {0.50}  &   	\bf{0.48}&\bf{	0.71}	&\bf{1.37}&	\bf{0.85}\\
\bottomrule
\end{tabular}}
\end{center}
\caption{Results on ShapeNet-55 in terms of  L2 Chamfer Distance $\times$ $10^3$ (lower is better). } \label{tab:shapenet55}
\end{table*}

\subsection{Loss Function}
We use Chamfer Distance (CD)~\cite{fan2017point} of the L1-norm as the primary component of our loss function.
We supervise all predicted point sets $\mathbb P = \{ {\mathcal P}_s, {\mathcal P_1}, {\mathcal P_2}, {\mathcal P_3}\}$ with ground truth ${\mathcal P}_t \in \mathbb R^{N_t \times 3}$, and the loss function $\mathcal L$ can be written as:

\begin{equation}
\mathcal L = \sum_{{\mathcal P}_g \in \mathbb P} {\rm CD}({\mathcal P}_g, {\rm FPS}({\mathcal P}_t, N_g)).
\end{equation}\label{eqt:cd}

{\noindent}Here, we use FPS algorithm~\cite{qi2017pointnet++} to down-sample ${\mathcal P}_t$ to the same density ({\em i.e.},  $N_g \times 3$) as with ${\mathcal P}_g$.

\section{Experiments}
In this section, we first conduct extensive experiments to demonstrate the effectiveness of our model on three widely used benchmark datasets including PCN~\cite{yuan2018pcn}, ShapeNet-55/34~\cite{yu2021pointr} and MVP~\cite{pan2021multi}. 
Then, we conduct the ablation studies on PCN dataset to analyze different designs of our model.

\subsection{Training Setup}
We implement CRA-PCN with Pytorch \cite{paszke2019pytorch}. 
All  models are trained on two NVIDIA Tesla V100 graphic cards with a batch size of 72. We use Adam  optimizer \cite{kingma2014adam} with $\beta_1 = 0.9$ and $\beta_2 = 0.999$. 
We set the feature dim $D$ in the decoder to 128.
The initial learning rate is set to 0.001 with continuous decay of 0.1 for every 100 epochs. 
We train our model for 300 epochs on PCN dataset \cite{yuan2018pcn}  and ShapeNet-55/34  \cite{yu2021pointr}  while 50 epochs on MVP dataset \cite{pan2021multi}.

\subsection{Evaluation on PCN Dataset}
\subsubsection{Dataset and metric.}
The PCN dataset~\cite{yuan2018pcn} is the subset of ShapeNet dataset~\cite{chang2015shapenet} and it consists of 30,974 shapes from 8 categories. 
Each ground truth point cloud is generated by evenly sampling 16,384 points on the mesh surface. 
The partial inputs of 2,048 points are generated by back-projecting 2.5D depth images into 3D. 
For a fair comparison, we follow the standard evaluation metric in~\cite{yuan2018pcn}, and all results on PCN dataset are evaluated in terms of L1 Chamfer Distance (CD-$\ell$1).

\subsubsection{Quantitative results.} 
In Tab.~\ref{tab:pcn1}, we report the quantitative results of our CRA-PCN and other methods on PCN dataset. 
We see that CRA-PCN achieves the best performance over all previous methods in all categories. 
Compared to the second-ranked AnchorFormer, our method reduces the average CD-$\ell$1 by $0.20$, which is $3.0\%$ lower than AnchorFormer. 
While several previous methods~\cite{xiang2021snowflakenet, zhou2022seedformer,yan2022fbnet} in Tab.~\ref{tab:pcn1} also use a similar coarse-to-fine architecture, CRA-PCN outperforms them by a large margin with the help of several novel designs.

\subsubsection{Qualitative results.} 
In Fig.~\ref{fig:pcn}, we visually compare CRA-PCN with previous state-of-the-art methods in four selected categories ({{\em Plane}, {\em Table}, {\em Lamp}, and {\em Chair}).
 The visual results show that CRA-PCN can predict smoother surfaces and produce less noise. 
Specifically, in \textit{Table} category in the second row, the visible seat predicted by CRA-PCN has  smooth details, while other methods generate many noise points. 
 As for \textit{Lamp} category in the third row, all other methods fail to reconstruct the missing spherical surface.

\begin{table*}[!t]
\begin{center}
\setlength{\tabcolsep}{9pt}{
\begin{tabular}{c |cccc | cccc}
\toprule
    	&\multicolumn{4}{c}{34 seen catagories}\vline &\multicolumn{4}{c}{21 unseen categories} \\
Methods 		& CD-S	& CD-M 	& CD-H 	&CD-Avg 	&CD-S 	&CD-M 	&CD-H 	&CD-Avg\\
\midrule
FoldingNet~\cite{yang2018foldingnet}	&1.86 	&1.81 	&3.38 	&2.35	&2.76	& 2.74	&5.36	&3.62\\
PCN~\cite{yuan2018pcn}			&1.87	&1.81	&2.97	&2.22	&3.17	&3.08	&5.29	&3.85\\
TopNet~\cite{tchapmi2019topnet}			&1.77	&1.61	&3.54	&2.31	&2.62	&2.43	&5.44	&3.50\\
PFNet~\cite{huang2020pf}		&3.16	&3.19	&7.71	&4.68	&5.29	&5.87	&13.33	&8.16\\	
GRNet~\cite{xie2020grnet}		&1.26	&1.39	&2.57	&1.74	&1.85	&2.25	&4.87	&2.99\\
PoinTr~\cite{yu2021pointr}		&0.76	&1.05	&1.88	&1.23	&1.04	&1.67	&3.44	&2.05\\
SeedFormer~\cite{zhou2022seedformer} 	&0.48	&0.70	&1.30	&0.83	&0.61	&1.07	&2.35	&1.34\\
\midrule
Ours &\bf{0.45}	&\bf{0.65}	&\bf{1.18}	&\bf{0.76}	&\bf{0.55}	&\bf{0.97}	&\bf{2.19}	&\bf{1.24}\\
\bottomrule
\end{tabular}}
\end{center}
\caption{Results on ShapeNet-34 in terms of  L2 Chamfer Distance $\times$ $10^3$ (lower is better). } \label{tab:shapenet34}
\end{table*}

\subsection{Evaluation on ShapeNet-55/34}

\subsubsection{Dataset and metric.}
To further evaluate the generalization ability of our model, we conduct experiments on ShapeNet-55 and ShapeNet-34~\cite{yu2021pointr}. Like PCN dataset, ShapeNet-55 and ShapeNet-34 are derived from ShapeNet, but contain all the objects in ShapeNet from 55 categories.
Specifically, ShapeNet-55 contains 41,952 models for training and 10,518 models for testing,
while the training set of ShapeNet-34 contains 46,765 objects from 34 categories and its testing set consists of 3,400 objects from 34 seen categories and 2,305 objects from the remaining 21 novel categories. Using the training and evaluation protocol proposed in~\cite{yu2021pointr}, partial inputs are generated online. 
Specifically, partial point clouds are generated by selecting certain viewpoints and removing $n$ percentage farthest points of complete shapes, where viewpoints are randomly selected during training and  fixed during evaluation. 
During the evaluation, $n$ is set to 25\%, 50\%, and 75\%, respectively, corresponding to three difficulty degrees, namely {\textit {simple}}, {\textit {moderate}} and {\textit {hard}}. 
Following~\cite{yu2021pointr}, we evaluate the performance of the methods~\cite{yang2018foldingnet, yuan2018pcn, tchapmi2019topnet, huang2020pf, xie2020grnet, yu2021pointr, zhou2022seedformer} in terms of CD-$\ell$2 under the above three difficulties, and we also report the average CD. 

\begin{table}
\begin{center}
\resizebox{1.0\linewidth}{!}{
\begin{tabular}{c|ccc}
\toprule

Methods  &CD-$\ell$2 $\downarrow$ &F-Score\%$\uparrow$ \\
\midrule
 PCN~\cite{yuan2018pcn} & 9.77 &0.320\\
 TopNet~\cite{tchapmi2019topnet} &10.11&0.308\\	
 MSN~\cite{liu2020morphing}	&7.90&0.432\\
 CDN~\cite{wang2020cascaded}	&7.25&0.434\\
 ECG~\cite{pan2020ecg}	&6.64&0.476\\
VRCNet~\cite{pan2021variational} &5.96&0.499\\	
\midrule
Ours&\bf{5.33} &\bf{0.529}\\
\bottomrule
\end{tabular}}
\end{center}
\caption{Results on MVP dataset in terms of  L2 Chamfer Distance $\times$ $10^4$ (lower is better) and F-Score (higher is better). } \label{tab:mvp}
\end{table}

\subsubsection{Results on ShapeNet-55.} 
The last four columns of Tab.~\ref{tab:shapenet55} show the superior performance of CRA-PCN under various situations with diverse viewpoints and diverse incomplete patterns.
Besides, we sample 10 categories and report CD-$\ell$2 of them.  
Categories in the columns from 2 to 6 contain sufficient samples, while the quantities of the following five categories are insufficient. 
According to the results, we can clearly see the powerful generalization ability of our model.

\subsubsection{Results on ShapeNet-34.} 
As shown in Tab.~\ref{tab:shapenet34}, our method achieves the best CD-$\ell$2 in the 34 seen categories of ShapeNet-34. 
Moreover, as for the 21 unseen categories that do not appear in training phase, 
our method outperforms other competitors. 
Especially, we achieve $6.8\%$ improvement compared with the second-ranked SeedFormer under hard setting in 21 unseen categories, which justifies the effectiveness of our method.

\begin{table}[!t]
\begin{center}
\resizebox{\linewidth}{!}{
\begin{tabular}{c|cc|c}
\toprule
Methods & Latency $\downarrow$ & Memory $\downarrow$ & CD $\times 10^3$ \\
\midrule
FBNet & 0.905 s & 3421 MB & 6.94 \\
SeedFormer & 0.297 s& 10701 MB& 6.74 \\
AnchorFormer & 0.608 s& 3735 MB& 6.59 \\
\midrule
Ours & 0.469 s& 4459 MB& 6.39 \\
\bottomrule
\end{tabular}}
\end{center}
\caption{Run-time memory usage and latency, which were evaluated on a single GTX 1080Ti graphic card with a batch size of 32.} 
\label{tab:let}
\end{table}

\begin{table}[!t]
\begin{center}
\resizebox{.9\linewidth}{!}{
\begin{tabular}{c|cc|c}
\toprule
Variations & Inter-level & Intra-level & CD-$\ell$1 \\
\midrule
A  &  &  & 7.85 \\
B & & \checkmark  & 6.92 \\
C  & \checkmark & & 6.80 \\
D & &$\times 2$ & 6.74 \\
E & $\times2$&  &6.61 \\
F & second & first  & 6.54\\
G  & \checkmark & \checkmark & \bf{6.39} \\
\bottomrule
\end{tabular}}
\end{center}
\caption{Analysis of  inter- and intra-level Cross-Resolution Transformer in up-sampling block.} \label{tab:ab1}
\end{table}

\begin{table}[!t]
\begin{center}
\resizebox{.9\linewidth}{!}{
\begin{tabular}{c|cc|c}
\toprule
Variations& inter-level & intra-level &CD-$\ell$1 \\
\midrule
A & $m=1$  & $m = 1$ & 6.75\\
B & $m=1$   & $m =3$ & 6.50\\
C & $m=3$  & $m =1$ & 6.51\\
D & $m=3$  & $m =3$ & \bf{6.39}\\
\bottomrule
\end{tabular}}
\end{center}
\caption{Analysis of the number of scales in Cross-Resolution Transformer.} 
\label{tab:ab3}
\end{table}

\subsection{Evaluation on MVP Dataset}
\subsubsection{Dataset and metric.}
The MVP dataset~\cite{pan2021multi} is a  multi-view partial point cloud dataset which consists of 16 categories of high-quality partial/complete point clouds. 
For each complete shape, 26 partial point clouds are generated by selecting 26 camera poses which are uniformly distributed on a unit sphere. MVP dataset provides various resolutions of ground truth, and we choose 2,048 among them.
In training phase, we follow the way of data splitting in~\cite{pan2021variational}. In evaluation phase, we evaluate performances in terms of CD-$\ell$2 and F-Score~\cite{tatarchenko2019single}.

\subsubsection{Results.}
We report the results in Tab.~\ref{tab:mvp} and our method outperforms all competitors.
Compared with VRCNet~\cite{pan2021variational}, our method reduces CD-$\ell$2 by $0.63$ and improves F-Score by $0.03$, which demonstrates that CRA-PCN not only predicts the complete shape but also preserves the details of partial input.

\subsection{Accuracy-Complexity Trade-Offs}
We report the run-time memory usage and latency in Tab.~\ref{tab:let}.
All methods were evaluated on a single Nvidia GeForce GTX 1080Ti graphic card with a batch size of 32.
For fair comparisons, we disable gradient calculation and use point clouds with a resolution of 2,048.
From the results, we see CRA-PCN can achieve better trade-offs than prior methods.

\subsection{Ablation Studies} \label{sec:ablation}
Here, we present ablation studies demonstrating the effectiveness of several proposed operations. 
All experiments are conducted under unified settings on the PCN dataset.

\subsubsection{Analysis of Cross-Resolution Transformers.} 
Tab.~\ref{tab:ab1} summarizes the evaluation results of inter- and intra-level Cross-Resolution Transformers, where model G is our CRA-PCN. 
We remove both inter- and intra-level CRT (model A), and we find the performance drops a lot, which justifies the effectiveness of cross-resolution aggregation.
Then we remove inter-level CRT (model B) or intra-level CRT (model C), respectively.
To keep the number of parameters the same with CRA-PCN, we double the number of intra-level CRT (model D) or inter-level CRT (model E). 
However, the aforementioned four variations exhibit poor performances compared to model G, and the results highlight the importance of the combination of intra- and inter-level cross-resolution aggregation. 
Moreover, we switch the order of inter- and intra-level CRT (model F), and we can find the performance drops a little.  
The empirical results here justify the effectiveness of Cross-Resolution Transformer and the combination of inter- and intra-level CRA.

\subsubsection{Analysis of recursive/multi-scale designs.} 
To justify the advantages of our recursive  designs on multiple scales  for point cloud completion, we conduct ablation study and report the results in Tab.~\ref{tab:ab3}. 
Model D is our CRA-PCN, and we set the number of scales $m$ to 3. 
Note that, if we set $m$ to 1, Cross-Resolution Transformer will degrade to point transformer~\cite{zhao2021point} which can only achieve single-scale cross-resolution  aggregation. 
We replace all multi-scale CRTs with single-scale ones (model A) by setting both $m$ to 1, and we can clearly find the performance drops a lot. 
We separately set $m$ of inter-level CRT and intra-level CRT to 1 (model B and  C), they increase CD-$\ell$1 by $0.11$ and $0.12$, respectively.    
The results of these three variations confirm the effectiveness of our recursive and multi-scale designs.

\section{Conclusion}
In this paper, focusing on explicit cross-resolution aggregation, we present Cross-Resolution Transformer that efficiently performs cross-resolution and  multi-scale feature aggregation. 
Moreover, we propose a combination of intra- and inter-level cross-resolution aggregation with the unified designs of Cross-Resolution Transformer.
Based on the aforementioned techniques, we propose a novel deep network for point cloud completion, named CRA-PCN, which adopts an encoder-decoder architecture.
Extensive experiments demonstrate the superiority of our method.
It will be an interesting future direction to extend our work to similar tasks such as point cloud reconstruction and up-sampling.

\section{Acknowledgments}
This work was supported by the National Natural Science Foundation of China (Grant No. 62372223).

\bibliography{aaai24}

\end{document}